\documentclass[10pt,twocolumn,letterpaper]{article}

\usepackage{cvpr}              

\usepackage[accsupp]{axessibility}
\usepackage{algorithm}
\usepackage{algpseudocode}
\usepackage{booktabs}
\usepackage{multirow}
\usepackage{rotating}
\usepackage{adjustbox}
\usepackage{bm}
\usepackage{soul}
\def\eg{\emph{e.g}\onedot}

\def\ie{\emph{i.e}\onedot}

\definecolor{cvprblue}{rgb}{0.21,0.49,0.74}
\usepackage[pagebackref,breaklinks,colorlinks,allcolors=cvprblue]{hyperref}

\def\paperID{11} 
\def\confName{CVPR}
\def\confYear{2025}

\title{MegaLoc: One Retrieval to Place Them All}

\author{Gabriele Berton\\
Polytechnic of Turin\\
{\tt\small bertongabri@gmail.com}
\and
Carlo Masone\\
Polytechnic of Turin, Focoos AI\\
}

\begin{document}
\maketitle

\begin{abstract}

Retrieving images from the same location as a given query is an important component of multiple computer vision tasks, like Visual Place Recognition, Landmark Retrieval, Visual Localization, 3D reconstruction, and SLAM.
However, existing solutions are built to specifically work for one of these tasks, and are known to fail when the requirements slightly change or when they meet out-of-distribution data.
In this paper we combine a variety of existing methods, training techniques, and datasets to train a retrieval model, called MegaLoc, 
that is performant on multiple tasks.
We find that MegaLoc
(1) achieves state of the art on a large number of Visual Place Recognition datasets,
(2) impressive results on common Landmark Retrieval datasets, 
and (3) sets a new state of the art for Visual Localization on the LaMAR datasets, where we only changed the retrieval method to LaMAR's official localization pipeline.
The code for MegaLoc is available at \url{https://github.com/gmberton/MegaLoc}

\end{abstract}

\section{Introduction}
\label{sec:introduction}

This paper tackles the task of retrieving images from a large database that represent the same place as a given query image.
But what does it mean for two images to be ``from the same place''?
Depending on who you ask, you'll get different answers:
\begin{enumerate}
    \item Landmark Retrieval (\textbf{LR}) folks will tell you that two photos are from the same place if they depict the same landmark, regardless of how close to each other the two photos were taken \cite{Weyand_2020_gldv2};
    \item Visual Place Recognition (\textbf{VPR}) people set a camera pose distance of 25 meters to define if two images are positives (\ie from the same place) \cite{Arandjelovic_2018_netvlad};
    \item Visual Localization (\textbf{VL}) / 3D Vision researchers will tell you that two images need to have their pose as close as possible to be considered the same place.
\end{enumerate}

\begin{figure}
    \begin{center}
    \includegraphics[width=0.99\columnwidth]{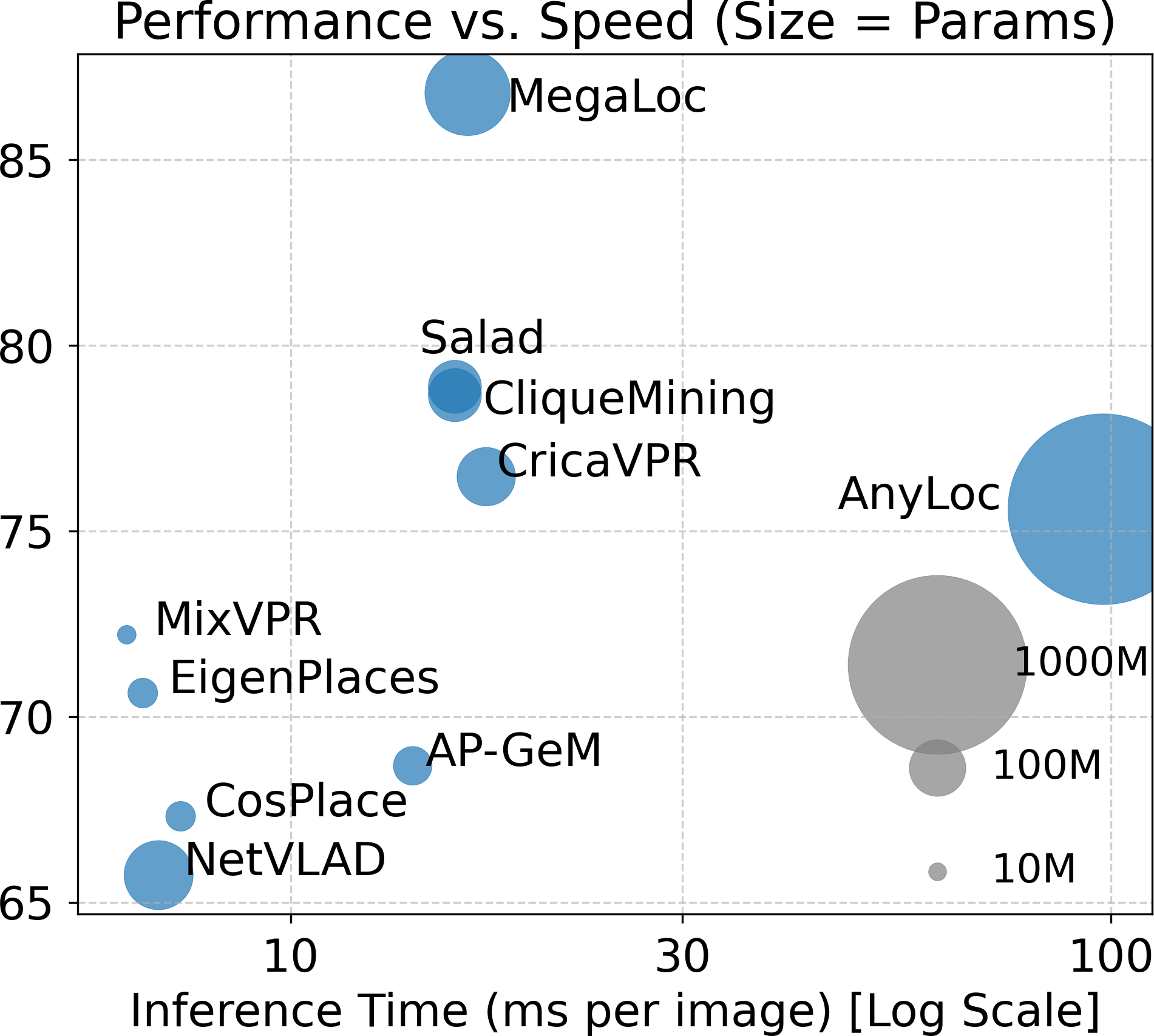}
    \end{center}
    \caption{Performance comparison of various Visual Place Recognition models.
    The y axis shows the results averaged across all evaluation datasets, the x axis shows the inference time, and the size of the circle represents the number of parameters per model.}
    \label{fig:plots}
\end{figure}

\begin{figure}
    \begin{center}
    \includegraphics[width=0.99\columnwidth]{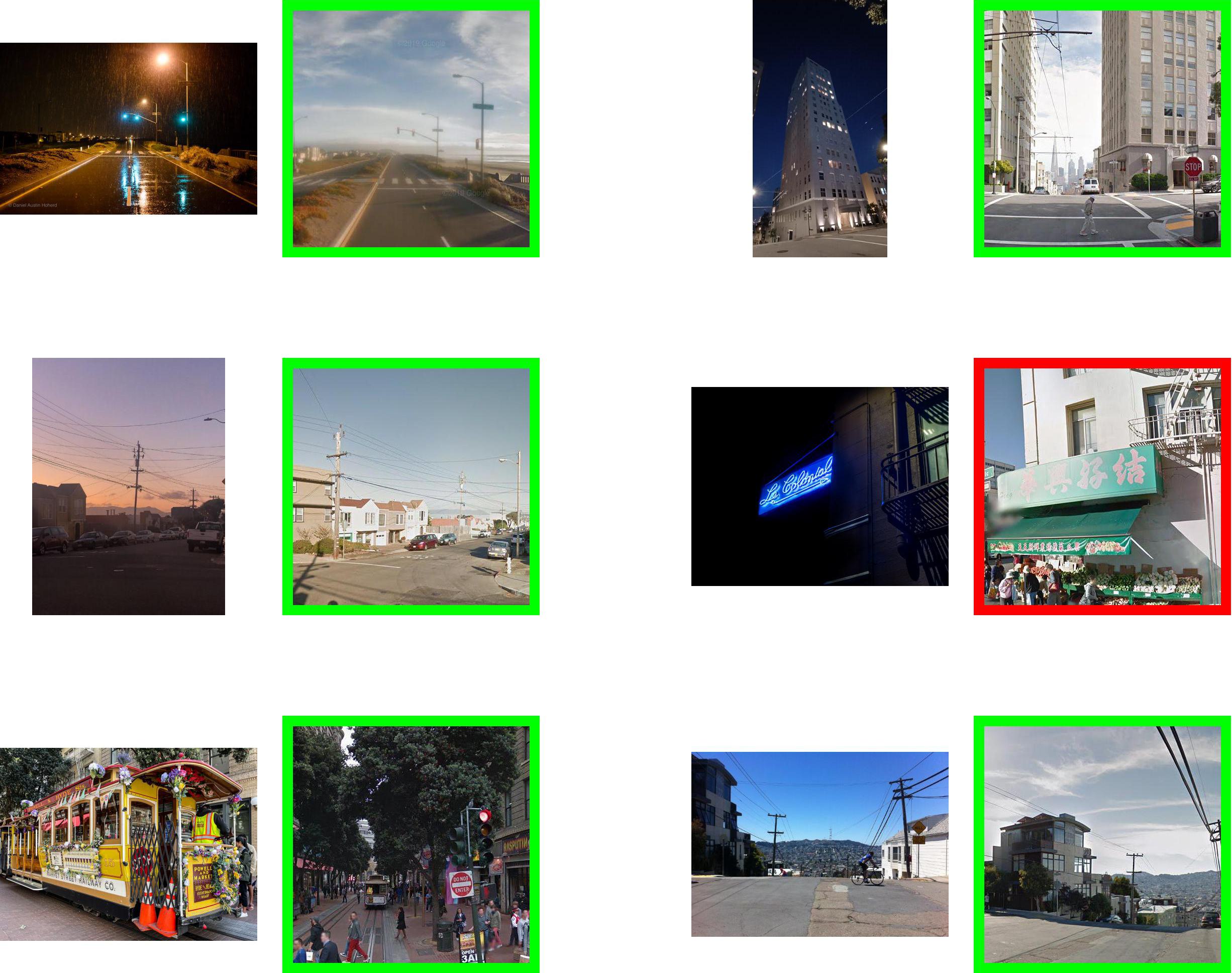}
    \end{center}
    \caption{Qualitative examples of predictions by MegaLoc. Each pair of images represents a query and its top-1 prediction from the SF-XL dataset, searched across the 2.8M database spanning 150 km$^2$ across San Francisco. Predictions in green are correct, red are wrong.}
    \label{fig:teaser}
\end{figure}

Even though image retrieval is a core component in all three tasks, their different definitions and requirement has inevitably led to the development of ad-hoc image retrieval solutions for each of them.
As these three tasks continued to diverge, over the years papers have avoided showing results of their methods on more than one of these tasks: VPR papers don't show results on LR, and LR papers don't show results on VPR.
In the meantime, 3D vision pipelines like COLMAP \cite{Schoenberger_2016_sfm_colmap}, Hierarchical Localization \cite{Sarlin_2019_hloc} and GLOMAP \cite{pan2024glomap} keep using outdated retrieval methods, like RootSIFT with bag-of-words \cite{Arandjelovic_2012_rootSift, Schonberger_2016_retrieval, Csurka_2003_bow} and NetVLAD \cite{Arandjelovic_2018_netvlad}.
In this paper we aim to put an end to this, by training a single model that achieves SOTA (or almost) on all of these tasks, showcasing robustness across diverse domains.
To train this model we do not propose any ``technical novelty'', but we use all the lessons learned from all these three task, putting together a combination of good samplers, datasets, and general training techniques.

``\emph{Why does it matter?}'', you may ask.
Imagine you are doing 3D reconstruction, where image retrieval is a fundamental component, on a collection of diverse scenes (\eg to create datasets like MegaDepth \cite{Li_2018_megadepth}, MegaScenes \cite{Tung_2024_Megascenes}, or for the evergreen Image Matching Challenge \cite{bellavia_2024_image}).
In some cases there would be small scenes (\eg reconstruction of a fountain), requiring a retrieval model that is able to retrieve nearby images (few meters away), which is something VPR models excel at, but LR models underperform (see \cite{Berton_2022_benchmark} Tab. 14).
In other cases however, the scene might be large (\eg a big landmark like a church), with images hundreds of meters away: while LR models are designed for this, VPR models achieve poor results in this situations (see \cref{sec:results_lr}).
Given these considerations, we note how neither VPR nor LR provide models for the diverse cases of 3D reconstructions, creating a gap in literature that is filled by MegaLoc.
As another example where a model like MegaLoc is necessary, one can think of Visual Place Recognition (which is also the first step for Visual Localization), where models are evaluated by using a 25 meters threshold (and queries in popular datasets always have at least one positive within 25 meters).
However, in the real world the nearest image to a given query might be 100 meters away, and while ideally we would still want to retrieve it, a VPR model is unlikely to work in such case, as it has been trained to ignore anything further away from the camera.

In this paper we demonstrate that, by leveraging a diverse set of data sources and best practices from LR, VPR and VL, we obtain a single image retrieval model that works well across all these tasks.

\section{Method}
\label{sec:method}

The core idea of this paper is to fuse data from multiple datasets, and train a single model.
We use five datasets containing both outdoor and indoor images (thorough description below) and catering to different image localization tasks: GSV-Cities \cite{Alibey_2022_gsvcities}, 
Mapillary Street-Level Sequences (MSLS) \cite{Warburg_2020_msls},
MegaScenes \cite{Tung_2024_Megascenes}, ScanNet \cite{Dai_2017_scannet} and San Francisco eXtra Large (SF-XL) \cite{Berton_2022_cosPlace}. 
At each training iteration, we extract six sub-batches of data, one for each dataset (except SF-XL, from which two sub-batches are sampled) and use a multi-similarity loss \cite{Wang_2019_multi_similarity_loss} computed over each sub-batch.
Each sub-batch is made of 128 images, containing 4 images (called quadruplets) from 32 different places/classes.
Given that these datasets have diverse format, they require different sampling techniques.
In the following paragraphs we explain how data is sampled from each dataset.

\paragraph{San Francisco eXtra Large (SF-XL)} is a dataset of 41M images with GPS and orientation from 12 different years, densely covering the entire city of San Francisco across time.
To select ideal quadruplets for training, we use the sampling technique presented in EigenPlaces \cite{Berton_2023_EigenPlaces}.
This method assures that each class contains images that represent a given place from diverse perspectives, while ensuring that no visual overlap exists between two different places.
EigenPlaces provides two sub-batches, one made of frontal-facing images (\ie with the camera facing straight along the street) and one of lateral-facing images.

\paragraph{Google Street View Cities (GSV-Cities)} is a dataset of 530k images split into 62k places/classes from 40 cities, where each class contains at least 4 images with same orientation and is at least 100 meters from any other class.
Given that GSV-Cities is already split into non-overlapping classes, it is not strictly necessary to apply a particular sampling technique.
We therefore directly feed the GSV-Cities dataset to the multi-similarity loss, as in the original GSV-Cities paper \cite{Alibey_2022_gsvcities}.

\paragraph{Mapillary Street-Level Sequences (MSLS)}
is a dataset of 1.6M images split in contiguous sequences, across 30 different cities over 9 years.
To ideally sample data from the MSLS dataset, we use the mining technique described in the CliqueMining paper \cite{Izquierdo_2024_cliqueM}.
This method ensures that the places selected for each batch depict visually similar (but geographically different) places (\ie hard negatives), so that the loss can be as high as possible and effectively teach the model to disambiguate between similar-looking places.

\paragraph{MegaScenes}
is a collection of 100k 3D structure-from-motion reconstructions, composed of 2M images from Wikimedia Commons.
Simply using each reconstruction as a class, and sampling random images from such class, could lead to images that do not have any visual overlap, \eg two images could show opposites facades of a building, therefore having no visual overlap while belonging to the same 3D reconstruction.
Therefore we make sure that when we sample a set of four images from a given reconstruction, each of these four images should have visual overlap with each other (we define visual overlap as having at least 1\% of 3D points in common in the 3D reconstruction).

\paragraph{ScanNet}
is a dataset of 2.5M views from 1500 scans from 707 indoor places.
To train on ScanNet we use each scene as a class, and select quadruplets so that each pair of images within a quadruplet has visual overlap (\ie less than 10 meters and 30° apart);
simultaneously we ensure that no two images from different quadruplets has visual overlap.

\section{Experiments}
\label{sec:experiments}

\begin{table*}
\begin{center}
\begin{adjustbox}{width=0.99\linewidth}
\centering
\begin{tabular}{lcccccccccccccccccccccccccccccccccccccccccccccccccccc}
\toprule
\multicolumn{1}{l}{\multirow{2}{*}{Method}} & Desc. & 
\multicolumn{2}{c}{Baidu \cite{Sun_2017_Baidu_dataset}} & &
\multicolumn{2}{c}{Eynsham \cite{Cummins_2009_eynsham, Berton_2022_benchmark}} & &
\multicolumn{2}{c}{MSLS val \cite{Warburg_2020_msls}} & &
\multicolumn{2}{c}{Pitts250k \cite{Gronat_2013_cvpr_pitts, Arandjelovic_2018_netvlad}} & &
\multicolumn{2}{c}{Pitts30k \cite{Gronat_2013_cvpr_pitts, Arandjelovic_2018_netvlad}} & &
\multicolumn{2}{c}{SF-XL v1 \cite{Berton_2022_cosPlace}} & &
\multicolumn{2}{c}{SF-XL v2 \cite{Berton_2022_cosPlace}} & &
\multicolumn{2}{c}{SF-XL night \cite{Barbarani_2023_local_features_benchmark}} & &
\multicolumn{2}{c}{SF-XL occlusion \cite{Barbarani_2023_local_features_benchmark}} & &
\multicolumn{2}{c}{Tokyo 24/7 \cite{Torii_2018_tokyo247}} \\
\cline{3-4} \cline{6-7} \cline{9-10} \cline{12-13} \cline{15-16} \cline{18-19} \cline{21-22} \cline{24-25} \cline{27-28} \cline{30-31}
& Dim.
& R1 & R10 & & R1 & R10 & & R1 & R10 & & R1 & R10 & & R1 & R10 & & R1 & R10 & & R1 & R10 & & R1 & R10 & & R1 & R10 & & R1 & R10 \\
\midrule
NetVLAD \cite{Arandjelovic_2018_netvlad}       &  4096 & 69.0&95.0 && 77.7&90.5 && 54.5&70.4 && 85.9&95.0 && 85.0&94.4 && 40.1&57.7 && 76.9&91.1 &&  6.7&14.2 &&  9.2&22.4 && 69.8&82.9 \\
AP-GeM \cite{Revaud_2019_ap_gem}       &  2048 & 59.8&90.8 && 68.3&84.0 && 56.0&72.9 && 80.0&93.5 && 80.7&94.1 && 37.9&54.1 && 66.4&84.6 &&  7.5&16.7 &&  5.3&14.5 && 57.5&77.5 \\
CosPlace \cite{Berton_2022_cosPlace}     &  2048 & 52.0&80.4 && 90.0&94.9 && 85.0&92.6 && 92.3&98.4 && 90.9&96.7 && 76.6&85.5 && 88.8&96.8 && 23.6&32.8 && 30.3&44.7 && 87.3&95.6 \\
MixVPR \cite{Alibey_2023_mixvpr}        &  4096 & 71.9&94.7 && 89.6&94.4 && 83.2&91.9 && 94.3&98.9 && 91.6&96.4 && 72.5&80.9 && 88.6&95.0 && 19.5&30.5 && 30.3&38.2 && 87.0&94.0 \\
EigenPlaces \cite{Berton_2023_EigenPlaces}   &  2048 & 69.1&91.9 && 90.7&95.4 && 85.9&93.1 && 94.1&98.7 && 92.5&97.6 && 84.0&90.7 && 90.8&96.7 && 23.6&34.5 && 32.9&52.6 && 93.0&97.5 \\
AnyLoc \cite{Keetha_2023_AnyLoc} & 49152 & \underline{75.6}&\underline{95.2} && 85.0&94.1 && 58.7&74.5 && 89.4&98.0 && 86.3&96.7 &&    -&   - &&    -&   - &&    -&   - &&    -&   - && 87.6&97.5 \\
Salad \cite{Izquierdo_2024_SALAD}         &  8448 & 72.7&93.6 && 91.6&95.9 && 88.2&95.0 && 95.0&\underline{99.2}&& 92.3&97.4 &&\underline{88.7}&\underline{94.4}&&\underline{94.6}&98.2&&\underline{46.1}&\underline{62.4}&&\underline{50.0}&\underline{68.4}&& 94.6&\underline{98.1} \\
CricaVPR \cite{Lu_2024_cricavpr}      & 10752 & 65.6&93.2 && 88.0&94.3 && 76.7&87.2 && 92.6&98.3&& 90.0&96.7 && 62.6&78.9 && 86.3&96.0 && 25.8&40.6 && 27.6&47.4 && 82.9&93.7 \\
CliqueMining \cite{Izquierdo_2024_cliqueM}  &  8448 & 72.9&92.7 &&\underline{91.9}&\underline{96.2} &&\textbf{91.6}&\textbf{95.9} &&\underline{95.3}&\underline{99.2}&&\underline{92.6}&\underline{97.8}&& 85.5&92.6 && 94.5&\underline{98.3}&& \underline{46.1}&60.9 && 44.7&64.5 &&\textbf{96.8}&97.8 \\
MegaLoc (Ours) & 8448 & \textbf{87.7}&\textbf{98.0}&&\textbf{92.6}&\textbf{96.8}&&\underline{91.0}&\underline{95.8}&&\textbf{96.4}&\textbf{99.3}&&\textbf{94.1}&\textbf{98.2}&&\textbf{95.3}&\textbf{98.0}&&\textbf{94.8}&\textbf{98.5}&&\textbf{52.8}&\textbf{73.8}&&\textbf{51.3}&\textbf{75.0}&&\underline{96.5}&\textbf{99.4} \\
\bottomrule
\end{tabular}
\end{adjustbox}
\end{center}
\caption{\textbf{Recall@1 and Recall@10 on multiple VPR datasets.} Best overall results on each dataset are in \textbf{bold}, second best results \underline{underlined}. Results marked with a ``-'' did not fit in 480GB of RAM (2.8M features of 49k dimensions require 560GB for a float32-based kNN).}
\label{tab:vpr}
\end{table*}

\begin{table*}
\begin{center}
\begin{adjustbox}{width=0.99\linewidth}
\centering
\begin{tabular}{lccccccccccccccccc}
\toprule
\multicolumn{1}{l}{\multirow{2}{*}{Method}} &
\multicolumn{2}{c}{CAB (Phone)} & &
\multicolumn{2}{c}{HGE (Phone)} & &
\multicolumn{2}{c}{LIN (Phone)} & &
\multicolumn{2}{c}{CAB (HoloLens)} & &
\multicolumn{2}{c}{HGE (HoloLens)} & &
\multicolumn{2}{c}{LIN (HoloLens)} \\
\cline{2-3} \cline{5-6} \cline{8-9} \cline{11-12} \cline{14-15} \cline{17-18}
& (1, 0.1) & (5, 1.0) & & (1, 0.1) & (5, 1.0) & & (1, 0.1) & (5, 1.0) & & (1, 0.1) & (5, 1.0) & & (1, 0.1) & (5, 1.0) & & (1, 0.1) & (5, 1.0) \\
\midrule
NetVLAD               & 43.4 & 54.0 && 54.8 & 80.0 && 74.4 & 87.8 && 63.1 & 81.4 && 57.9 & 71.6 && 76.1 & 83.0 \\
AP-GeM                & 39.4 & 52.0 && 58.0 & 81.3 && 69.1 & 82.0 && 62.9 & 82.5 && 65.6 & 76.6 && 80.7 & 91.1 \\
Fusion (NetVLAD+AP-GeM)&41.4 & 53.8 && 56.3 & 82.4 && 76.0 & 89.4 && 63.2 & 83.1 && 63.1 & 75.1 && 78.5 & 87.0 \\
CosPlace              & 29.0 & 37.4 && 54.4 & 81.3 && 63.3 & 75.7 && 56.4 & 77.8 && 55.6 & 69.8 && 80.6 & 91.4 \\
MixVPR                & 40.9 & 50.8 && 59.2 & 83.8 && 77.5 & 89.8 && 65.2 & 84.7 && 63.3 & 74.7 && 83.6 & 92.2 \\
EigenPlaces           & 32.3 & 44.7 && 56.3 & 81.3 && 70.2 & 82.6 && 63.9 & 81.8 && 60.2 & 72.5 && 84.8 & 93.1 \\
AnyLoc                &\textbf{48.0}&\underline{59.8}&& 58.8 & 83.0 && 77.2 & 92.4 && 69.7 & 88.5 && 70.1 & 81.0 && 81.4 & 90.4 \\
Salad                 & 44.2 & 55.6 &&65.3&\underline{92.2}&&\underline{81.7}&\underline{94.0}&& 71.5 & 90.7 &&\underline{75.3}&\underline{85.2}&& 91.3 &\textbf{99.4}\\
CricaVPR              & 40.4 & 52.0 && 63.7 & 89.3 && 80.7 & 93.1 && 73.9 & 90.7 && 72.5 & 81.6 && 89.1 & 98.4 \\
CliqueMining          & 44.2 & 55.6 &&\underline{66.0}& 91.4 && 80.5 & 93.1 &&\underline{74.2}&\underline{90.9}&&\textbf{77.3}&\textbf{86.3}&&\underline{92.0}&98.8 \\
MegaLoc (Ours)        &\underline{47.0}&\textbf{60.4}&&\textbf{67.2}&\textbf{92.9}&&\textbf{83.3}&\textbf{94.9}&&\textbf{77.4}&\textbf{93.4}&&72.9&83.5&&\textbf{92.2}&\underline{99.0}\\
\bottomrule
\end{tabular}
\end{adjustbox}
\end{center}
\caption{\textbf{Results on LaMAR's datasets}, computed on each of the three locations, for both types of queries (HoloLens and Phone), which include both indoor and outdoor. For each location we report the recall at (1°, 10cm) and (5°, 1m), following the LaMAR paper \cite{sarlin2022lamar}.}
\label{tab:lamar}
\end{table*}

\subsection{Implementation details}
\label{sec:implementation_details}
During training, images are resized to 224$\times$224, while for inference we resize them to 322$\times$322, following \cite{Izquierdo_2024_SALAD}.
We use RandAugment \cite{Cubuk_2020_RandAugment} for data augmentation, as in \cite{Alibey_2022_gsvcities}, and AdamW \cite{Loshchilov_2018_AdamW} as optimizer.
Training is performed for 40k iterations.
The loss is simply computed as
$\mathcal{L} = \mathcal{L}_1 + \mathcal{L}_2 + \mathcal{L}_3 + \mathcal{L}_4 + \mathcal{L}_5 + \mathcal{L}_6$, where each $\mathcal{L}_n$ is the multi-similarity loss computed on one of the sub-batches.

\paragraph{The architecture} consists of a DINO-v2-base backbone \cite{Oquab_2023_dinov2} followed by a SALAD \cite{Izquierdo_2024_SALAD} aggregation layer, which has shown state-of-the-art performances over multiple VPR datasets \cite{Izquierdo_2024_SALAD, Izquierdo_2024_cliqueM}.
The SALAD layer is computed with 64 clusters, 256
channels per cluster, a global token of 256 and an MLP dimension of 512. The SALAD layer is followed by a linear projection (from a dimension of 16640 to 8448) and an L2 normalization.
During training, most of the model is frozen, except for the 4 last transformer layers, the SALAD layer and the final linear layer.

\paragraph{Memory-efficient GPU training} is achieved using PyTorch \cite{Paszke_2019_PyTorch}, where instead of summing all the losses and computing a single backward pass, we compute backward passes independently on each dataset: the backward pass computes and accumulates the gradient while freeing memory (see \cref{alg:mem_efficient_train}).
This simple technique reduces the VRAM for training MegaLoc from 360GB to 60GB (a 6x reduction because there are 6 losses) with \textbf{no increase in latency}, unlike in standard gradient accumulation where a batch is split into micro batches: this allows MegaLoc to be trained on a single A100 GPU.

\begin{algorithm}
\caption{Memory-Efficient GPU Training}
\label{alg:mem_efficient_train}
\begin{algorithmic}[1]
    \Require Model $Model$, Optim. $Opt$, Datasets $\{D_i\}_{i=1}^N$
    \State Initialize $Model$, $Opt$
    \For{each training iteration}
        \For{each dataset $D_i$ in $\{D_1, ..., D_N\}$}
            \State $B_i \leftarrow \text{LoadBatch}(D_i)$
            \State $Y_i \leftarrow Model(B_i)$
            \State $L_i \leftarrow \text{Loss}(Y_i, \text{target}_i)$
            \State $L_i.backward()$
        \EndFor
        \State $Opt.step()$
        \State $Opt.zero\_grad()$
    \EndFor
\end{algorithmic}
\end{algorithm}

\subsection{Results}
\label{sec:results}
We perform experiments on three different types of tasks:
\begin{itemize}
    \item Visual Place Recognition, where the task is to retrieve images that are within 25 meters from the query (\cref{sec:results_vpr});
    \item Visual Localization, where retrieval is part of a bigger pipeline that aims at finding the precise pose of the query given a set of posed images (\cref{sec:results_lamar});
    \item Landmark Retrieval, \ie retrieving images that depict the same landmark as the query (\cref{sec:results_lr}).
\end{itemize}

\subsubsection{Visual Place Recognition}
\label{sec:results_vpr}
We run experiments on a comprehensive set of Visual Place Recognition datasets.
These datasets contain a large variety of domains, including:
outdoor, indoor, street-view, hand-held camera, car-mounted camera, night, occlusions, long-term changes, grayscale.
Results are shown in \cref{tab:vpr}.
While other high-performing VPR models (like SALAD and CliqueMining) achieve very good results (\ie comparable to MegaLoc) on most datasets, MegaLoc vastly outperforms every other model on Baidu, which is an indoor-only dataset.

\begin{table}
\begin{center}
\begin{adjustbox}{width=0.9\linewidth}
\centering
\begin{tabular}{lccccccc}
\toprule
\multicolumn{1}{l}{\multirow{2}{*}{Method}} & \multicolumn{3}{c}{R-Oxford} & & \multicolumn{3}{c}{R-Paris}\\
\cline{2-4} \cline{6-8}
& E & M & H & & E & M & H\\
\hline
NetVLAD        & 24.1 & 16.1 &  4.7 && 61.2 & 46.3 & 22.0 \\
AP-GeM         & 49.6 & 37.6 & 19.3 && 82.5 & 69.5 & 45.5 \\
CosPlace       & 32.1 & 23.4 & 10.3 && 57.6 & 45.0 & 22.3 \\
MixVPR         & 38.2 & 28.4 & 10.8 && 61.9 & 48.3 & 25.0 \\
EigenPlaces    & 29.4 & 22.9 & 11.8 && 60.9 & 47.3 & 23.6 \\
AnyLoc  &\underline{64.2}&\underline{45.5}& 18.9 &&\underline{82.8}& 68.5 & 48.8 \\
Salad          & 55.2 & 42.3 & 21.4 && 76.6 & 66.2 & 44.8 \\
CricaVPR       & 57.0 & 39.2 & 15.3 && 80.0 & \underline{68.9} & \underline{48.9} \\
CliqueMining   & 52.2 & 41.0 &\underline{22.1}&& 71.8 & 60.5 & 41.2 \\
MegaLoc (Ours)&\textbf{91.0}&\textbf{79.0}&\textbf{62.1}&&\textbf{95.3}&\textbf{89.6}&\textbf{77.1}\\
\bottomrule
\end{tabular}
\end{adjustbox}
\end{center}
\caption{\textbf{Results on Landmark Retrieval datasets}, respectively Revisited Paris 6k \cite{Radenovic_CVPR_2018_roxford_rparis, Philbin_2008_paris6k} and Revisited Oxford 5k \cite{Radenovic_CVPR_2018_roxford_rparis, Philbin_2007_oxford5k}.}
\label{tab:landmark_retrieval}
\end{table}

\begin{figure*}
    \begin{center}
    \includegraphics[width=0.99\linewidth]{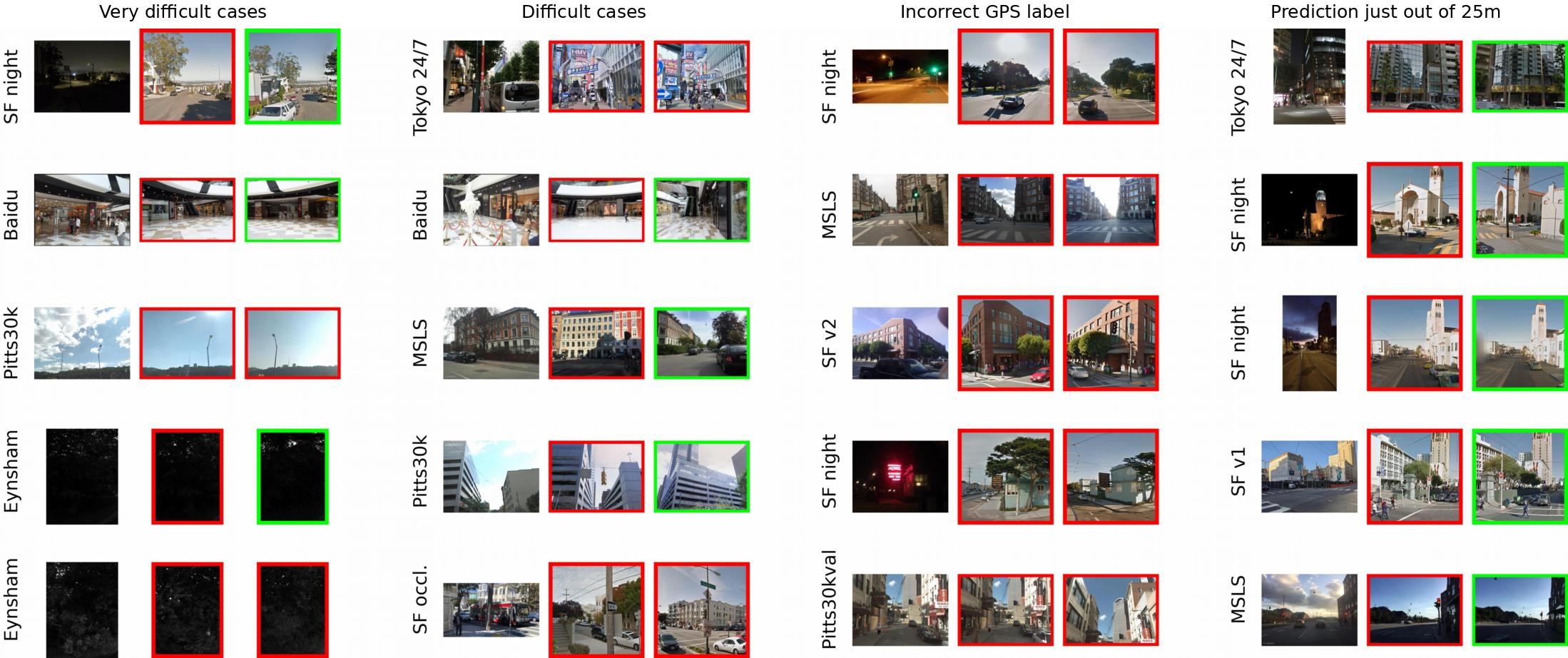}
    \end{center}
    \caption{\textbf{Failure cases, grouped in 4 categories.} Each one of the 4 column represent a category of failure cases: for each category we show 5 examples, made of 3 images, namely the query and its top-2 predictions with MegaLoc, which can be in red or green depending if the prediction is correct (\ie within 25 meters).
    The 4 categories that we identified are (1) \textit{very difficult cases}, which are unlikely to be solved any time soon; (2) \textit{difficult cases}, which can probably be solved by slightly better models than the current ones or simple post-processing; (3) \textit{incorrect GPS labels}, which, surprisingly, exist also in Mapillary and Google StreetView data; (4) \textit{predictions just out of the 25m threshold}, which despite being considered negatives in VPR, are actually useful predictions for real-world applications.}
    \label{fig:failure_cases}
\end{figure*}

\subsubsection{Visual Localization}
\label{sec:results_lamar}
Image retrieval is a core tool to solve 3D vision tasks, in pipelines like visual localization (\eg Hierarchical Localization \cite{Sarlin_2019_hloc} and InLoc \cite{Taira_2018_inloc}) and 3D reconstructions (\eg COLMAP \cite{Schoenberger_2016_mvs_colmap, Schoenberger_2016_sfm_colmap} and GLOMAP \cite{pan2024glomap}).
To understand if our method can help this use case, we compute results on both query sets (phone and HoloLens) of the three datasets of LaMAR \cite{sarlin2022lamar}, which comprise various challenges, including plenty of visual aliasing from both indoor and outdoor imagery.
To do this, we relied on the official LaMAR codebase\footnote{\url{https://github.com/microsoft/lamar-benchmark}} by simply replacing the retrieval method.
We computed these experiments on the validation set, as the labels for the test set are not publicly available.
Note that we only use LaMAR (or the landmark retrieval datasets) as a test set, and using it as validation would take unfeasibly long (multiple days).
Results are reported in \cref{tab:lamar}, and show that while some other models achieve good results on some datasets, MegaLoc is the only one that consistently achieves high results.

\subsubsection{Landmark Retrieval}
\label{sec:results_lr}
For the task of Landmark Retrieval we compute results on the most used datasets in literature, namely (the revisited versions of \cite{Radenovic_CVPR_2018_roxford_rparis}) Oxford5k \cite{Philbin_2007_oxford5k} and Paris6k \cite{Philbin_2008_paris6k}.
To do this we relied on the official codebase for the datasets\footnote{\url{https://github.com/filipradenovic/revisitop}}, by simply swapping the retrieval method.
Results, reported in \cref{tab:landmark_retrieval}, show a large gap between MegaLoc and previous VPR models on this task, which can be simply explained by the fact that previous models were only optimized for the standard VPR metric of retrieving images within 25 meters from the query.

\subsubsection{Failure Cases}
We identified a series of 4 main categories of ``failure cases'' that prevent the results from reaching 100\% recalls, and we present them in \cref{fig:failure_cases}.
We note however that, from a practical perspective, the only real failure cases are depicted in the second category/column of \cref{fig:failure_cases}: furthermore, in most similar cases SOTA models (\ie not only MegaLoc, but also other recent ones) can actually retrieve precise predictions, meaning that these failure cases can be likely solvable by some simple post-processing techniques (\eg re-ranking with image matchers, or majority voting).
Finally, another failure case that we noted, is when database images do not cover properly the search area: this is very common in the Mapillary (MSLS) dataset, where database images only show one direction (\eg photos along a road taken from north to south), while the queries are photos facing the other direction. We note however, that in the real world this can be easily solved by collecting database images in multiple directions, which is also common in most test datasets, like Eynsham, Pitts30k, Tokyo 24/7 and SF-XL.

\section{Conclusion and limitations}
\textit{So, is image retrieval for localization solved?}
Well, almost. While some datasets still show some room for improvement, we note that this is often due to either arguably unsolvable failure cases, wrong labels, and very few cases that can be solved by better models.
We emphasize however that this has been the case for some time, as previous DINO-v2-based models, like SALAD and CliqueMining, show very high results on classic VPR datasets.
What is still missing from literature is models like MegaLoc that achieve good results in a variety of diverse tasks and domains.

\textit{Should you always use MegaLoc?}
Well, almost, except for at least 3 use-cases.
MegaLoc has shown great results on a variety of related tasks, and, unlike other VPR models, achieves good results on landmark retrieval, which make it a great option also for retrieval for 3D reconstruction tasks, besides standard VPR and visual localization tasks.
However, experiments show that MegaLoc is outperformed by CliqueMining in MSLS, which is a dataset made of (almost entirely) forward facing images (\ie photos where the camera is facing the same direction of the street, instead of facing sideways towards the side of the street).
Another use case where MegaLoc is likely to be suboptimal is in very unusual natural environments, like forests or caves, where instead AnyLoc has been shown to work well \cite{Keetha_2023_AnyLoc}.
A third and final use case where other models might be preferred to MegaLoc is for embedded systems, where one might opt for more lightweight models, like the ResNet-18 \cite{He_2016_resnet} versions of CosPlace \cite{Berton_2022_cosPlace}, which has 11M parameters instead of MegaLoc's 228M.

\paragraph{Acknowledgements}
\noindent Carlo Masone was supported by FAIR - Future Artificial Intelligence Research which received funding from the European Union Next-GenerationEU (Piano nazionale di ripresa e resilienza (PNRR) – missione 4 componente 2, investimento 1.3 – D.D. 1555 11/10/2022, PE00000013).  This manuscript reflects only the authors’ views and opinions, neither the European Union nor the European Commission can be considered responsible for them.

\noindent We acknowledge the CINECA award
under the ISCRA initiative, for the availability of high performance computing resources.

{
    \small
    \bibliographystyle{ieeenat_fullname}
    \bibliography{main}

\begin{thebibliography}{40}
\providecommand{\natexlab}[1]{#1}
\providecommand{\url}[1]{\texttt{#1}}
\expandafter\ifx\csname urlstyle\endcsname\relax
  \providecommand{\doi}[1]{doi: #1}\else
  \providecommand{\doi}{doi: \begingroup \urlstyle{rm}\Url}\fi

\bibitem[Ali-bey et~al.(2022)Ali-bey, Chaib-draa, and Gigu{\`e}re]{Alibey_2022_gsvcities}
Amar Ali-bey, Brahim Chaib-draa, and Philippe Gigu{\`e}re.
\newblock Gsv-cities: Toward appropriate supervised visual place recognition.
\newblock \emph{Neurocomputing}, 513:\penalty0 194--203, 2022.

\bibitem[Ali-bey et~al.(2023)Ali-bey, Chaib-draa, and Gigu{\`e}re]{Alibey_2023_mixvpr}
Amar Ali-bey, Brahim Chaib-draa, and Philippe Gigu{\`e}re.
\newblock Mixvpr: Feature mixing for visual place recognition.
\newblock In \emph{Proceedings of the IEEE/CVF Winter Conference on Applications of Computer Vision}, pages 2998--3007, 2023.

\bibitem[Arandjelovic(2012)]{Arandjelovic_2012_rootSift}
Relja Arandjelovic.
\newblock Three things everyone should know to improve object retrieval.
\newblock In \emph{Proceedings of the 2012 IEEE Conference on Computer Vision and Pattern Recognition (CVPR)}, page 2911–2918, USA, 2012. IEEE Computer Society.

\bibitem[{Arandjelović} et~al.(2018){Arandjelović}, Gronat, Torii, Pajdla, and Sivic]{Arandjelovic_2018_netvlad}
Relja {Arandjelović}, Petr Gronat, Akihiko Torii, Tomas Pajdla, and Josef Sivic.
\newblock {NetVLAD}: {CNN} architecture for weakly supervised place recognition.
\newblock \emph{IEEE Transactions on Pattern Analysis and Machine Intelligence}, 40\penalty0 (6):\penalty0 1437--1451, 2018.

\bibitem[Barbarani et~al.(2023)Barbarani, Mostafa, Bayramov, Trivigno, Berton, Masone, and Caputo]{Barbarani_2023_local_features_benchmark}
Giovanni Barbarani, Mohamad Mostafa, Hajali Bayramov, Gabriele Trivigno, Gabriele Berton, Carlo Masone, and Barbara Caputo.
\newblock Are local features all you need for cross-domain visual place recognition?
\newblock In \emph{CVPRW}, pages 6155--6165, 2023.

\bibitem[Bellavia et~al.(2024)Bellavia, Matas, Mishkin, Morelli, Remondino, Sun, Tabb, Trulls, Yi, Dane, and Chow]{bellavia_2024_image}
Fabio Bellavia, Jiri Matas, Dmytro Mishkin, Luca Morelli, Fabio Remondino, Weiwei Sun, Amy Tabb, Eduard Trulls, Kwang~Moo Yi, Sohier Dane, and Ashley Chow.
\newblock Image matching challenge 2024 - hexathlon.
\newblock \url{https://kaggle.com/competitions/image-matching-challenge-2024}, 2024.
\newblock Kaggle.

\bibitem[Berton et~al.(2022)Berton, Masone, and Caputo]{Berton_2022_cosPlace}
Gabriele Berton, Carlo Masone, and Barbara Caputo.
\newblock Rethinking visual geo-localization for large-scale applications.
\newblock In \emph{IEEE Conference on Computer Vision and Pattern Recognition}, pages 4868--4878, 2022.

\bibitem[Berton et~al.(2023{\natexlab{a}})Berton, Mereu, Trivigno, Masone, Csurka, Sattler, and Caputo]{Berton_2022_benchmark}
Gabriele Berton, Riccardo Mereu, Gabriele Trivigno, Carlo Masone, Gabriela Csurka, Torsten Sattler, and Barbara Caputo.
\newblock Deep visual geo-localization benchmark, 2023{\natexlab{a}}.

\bibitem[Berton et~al.(2023{\natexlab{b}})Berton, Trivigno, Caputo, and Masone]{Berton_2023_EigenPlaces}
Gabriele Berton, Gabriele Trivigno, Barbara Caputo, and Carlo Masone.
\newblock Eigenplaces: Training viewpoint robust models for visual place recognition.
\newblock In \emph{Proceedings of the IEEE/CVF International Conference on Computer Vision (ICCV)}, pages 11080--11090, 2023{\natexlab{b}}.

\bibitem[Csurka et~al.(2004)Csurka, Dance, Fan, Willamowski, and Bray]{Csurka_2003_bow}
Gabriela Csurka, Christopher Dance, Lixin Fan, Jutta Willamowski, and Cédric Bray.
\newblock Visual categorization with bags of keypoints.
\newblock In \emph{European Conference on Computer Vision}, 2004.

\bibitem[Cubuk et~al.(2020)Cubuk, Zoph, Shlens, and Le]{Cubuk_2020_RandAugment}
Ekin~Dogus Cubuk, Barret Zoph, Jon Shlens, and Quoc Le.
\newblock Randaugment: Practical automated data augmentation with a reduced search space.
\newblock In \emph{Advances in Neural Information Processing Systems}, pages 18613--18624. Curran Associates, Inc., 2020.

\bibitem[Cummins and Newman(2009)]{Cummins_2009_eynsham}
M. Cummins and P. Newman.
\newblock Highly scalable appearance-only slam - {FAB-MAP} 2.0.
\newblock In \emph{Robotics: Science and Systems}, 2009.

\bibitem[Dai et~al.(2017)Dai, Chang, Savva, Halber, Funkhouser, and Nie{\ss}ner]{Dai_2017_scannet}
Angela Dai, Angel~X. Chang, Manolis Savva, Maciej Halber, Thomas Funkhouser, and Matthias Nie{\ss}ner.
\newblock Scannet: Richly-annotated 3d reconstructions of indoor scenes.
\newblock In \emph{Proc. Computer Vision and Pattern Recognition (CVPR), IEEE}, 2017.

\bibitem[Gronát et~al.(2013)Gronát, Obozinski, Sivic, and Pajdla]{Gronat_2013_cvpr_pitts}
Petr Gronát, Guillaume Obozinski, Josef Sivic, and Tomá Pajdla.
\newblock Learning and calibrating per-location classifiers for visual place recognition.
\newblock In \emph{2013 IEEE Conference on Computer Vision and Pattern Recognition}, pages 907--914, 2013.

\bibitem[{He} et~al.(2016){He}, {Zhang}, {Ren}, and {Sun}]{He_2016_resnet}
K. {He}, X. {Zhang}, S. {Ren}, and J. {Sun}.
\newblock Deep residual learning for image recognition.
\newblock In \emph{IEEE Conference on Computer Vision and Pattern Recognition}, pages 770--778, 2016.

\bibitem[Izquierdo and Civera(2024)]{Izquierdo_2024_SALAD}
Sergio Izquierdo and Javier Civera.
\newblock Optimal transport aggregation for visual place recognition.
\newblock In \emph{IEEE Conference on Computer Vision and Pattern Recognition}, 2024.

\bibitem[Keetha et~al.(2023)Keetha, Mishra, Karhade, Jatavallabhula, Scherer, Krishna, and Garg]{Keetha_2023_AnyLoc}
Nikhil Keetha, Avneesh Mishra, Jay Karhade, Krishna~Murthy Jatavallabhula, Sebastian Scherer, Madhava Krishna, and Sourav Garg.
\newblock Anyloc: Towards universal visual place recognition.
\newblock \emph{arXiv}, 2023.

\bibitem[Li and Snavely(2018)]{Li_2018_megadepth}
Zhengqi Li and Noah Snavely.
\newblock Megadepth: Learning single-view depth prediction from internet photos.
\newblock In \emph{Proceedings of the IEEE conference on computer vision and pattern recognition}, pages 2041--2050, 2018.

\bibitem[Loshchilov and Hutter(2019)]{Loshchilov_2018_AdamW}
Ilya Loshchilov and Frank Hutter.
\newblock Decoupled weight decay regularization.
\newblock In \emph{International Conference on Learning Representations}, 2019.

\bibitem[Lu et~al.(2024)Lu, Lan, Zhang, Jiang, Wang, and Yuan]{Lu_2024_cricavpr}
Feng Lu, Xiangyuan Lan, Lijun Zhang, Dongmei Jiang, Yaowei Wang, and Chun Yuan.
\newblock Cricavpr: Cross-image correlation-aware representation learning for visual place recognition.
\newblock In \emph{Proceedings of the IEEE/CVF Conference on Computer Vision and Pattern Recognition (CVPR)}, 2024.

\bibitem[Oquab et~al.(2023)Oquab, Darcet, Moutakanni, Vo, Szafraniec, Khalidov, Fernandez, Haziza, Massa, El-Nouby, Howes, Huang, Xu, Sharma, Li, Galuba, Rabbat, Assran, Ballas, Synnaeve, Misra, Jegou, Mairal, Labatut, Joulin, and Bojanowski]{Oquab_2023_dinov2}
Maxime Oquab, Timothée Darcet, Theo Moutakanni, Huy~V. Vo, Marc Szafraniec, Vasil Khalidov, Pierre Fernandez, Daniel Haziza, Francisco Massa, Alaaeldin El-Nouby, Russell Howes, Po-Yao Huang, Hu Xu, Vasu Sharma, Shang-Wen Li, Wojciech Galuba, Mike Rabbat, Mido Assran, Nicolas Ballas, Gabriel Synnaeve, Ishan Misra, Herve Jegou, Julien Mairal, Patrick Labatut, Armand Joulin, and Piotr Bojanowski.
\newblock Dinov2: Learning robust visual features without supervision, 2023.

\bibitem[Pan et~al.(2024)Pan, Barath, Pollefeys, and Sch\"{o}nberger]{pan2024glomap}
Linfei Pan, Daniel Barath, Marc Pollefeys, and Johannes~Lutz Sch\"{o}nberger.
\newblock {Global Structure-from-Motion Revisited}.
\newblock In \emph{European Conference on Computer Vision (ECCV)}, 2024.

\bibitem[Paszke et~al.(2019)Paszke, Gross, Massa, Lerer, Bradbury, Chanan, Killeen, Lin, Gimelshein, Antiga, Desmaison, Kopf, Yang, DeVito, Raison, Tejani, Chilamkurthy, Steiner, Fang, Bai, and Chintala]{Paszke_2019_PyTorch}
Adam Paszke, Sam Gross, Francisco Massa, Adam Lerer, James Bradbury, Gregory Chanan, Trevor Killeen, Zeming Lin, Natalia Gimelshein, Luca Antiga, Alban Desmaison, Andreas Kopf, Edward Yang, Zachary DeVito, Martin Raison, Alykhan Tejani, Sasank Chilamkurthy, Benoit Steiner, Lu Fang, Junjie Bai, and Soumith Chintala.
\newblock Pytorch: An imperative style, high-performance deep learning library.
\newblock In \emph{Advances in Neural Information Processing Systems 32}, pages 8024--8035. Curran Associates, Inc., 2019.

\bibitem[Philbin et~al.(2007)Philbin, Chum, Isard, Sivic, and Zisserman]{Philbin_2007_oxford5k}
James Philbin, Ondrej Chum, Michael Isard, Josef Sivic, and Andrew Zisserman.
\newblock Object retrieval with large vocabularies and fast spatial matching.
\newblock In \emph{IEEE Conference on Computer Vision and Pattern Recognition}. IEEE Computer Society, 2007.

\bibitem[Philbin et~al.(2008)Philbin, Chum, Isard, Sivic, and Zisserman]{Philbin_2008_paris6k}
James Philbin, Ondrej Chum, Michael Isard, Josef Sivic, and Andrew Zisserman.
\newblock Lost in quantization: Improving particular object retrieval in large scale image databases.
\newblock In \emph{IEEE Conference on Computer Vision and Pattern Recognition}, 2008.

\bibitem[Radenovi\'{c} et~al.(2018)Radenovi\'{c}, Iscen, Tolias, Avrithis, and Chum]{Radenovic_CVPR_2018_roxford_rparis}
F. Radenovi\'{c}, A. Iscen, G. Tolias, Y. Avrithis, and O. Chum.
\newblock Revisiting oxford and paris: Large-scale image retrieval benchmarking.
\newblock In \emph{CVPR}, 2018.

\bibitem[Revaud et~al.(2019)Revaud, Almaz{\'a}n, Rezende, and de~Souza]{Revaud_2019_ap_gem}
J{\'e}r{\^o}me Revaud, Jon Almaz{\'a}n, R.~S. Rezende, and C{\'e}sar~Roberto de Souza.
\newblock Learning with average precision: Training image retrieval with a listwise loss.
\newblock \emph{2019 IEEE/CVF International Conference on Computer Vision (ICCV)}, pages 5106--5115, 2019.

\bibitem[Sarlin et~al.(2019)Sarlin, Cadena, Siegwart, and Dymczyk]{Sarlin_2019_hloc}
Paul-Edouard Sarlin, Cesar Cadena, Roland Siegwart, and Marcin Dymczyk.
\newblock From coarse to fine: Robust hierarchical localization at large scale.
\newblock In \emph{CVPR}, 2019.

\bibitem[Sarlin et~al.(2022)Sarlin, Dusmanu, Schönberger, Speciale, Gruber, Larsson, Miksik, and Pollefeys]{sarlin2022lamar}
Paul-Edouard Sarlin, Mihai Dusmanu, Johannes~L. Schönberger, Pablo Speciale, Lukas Gruber, Viktor Larsson, Ondrej Miksik, and Marc Pollefeys.
\newblock {LaMAR: Benchmarking Localization and Mapping for Augmented Reality}.
\newblock In \emph{ECCV}, 2022.

\bibitem[Sch\"{o}nberger and Frahm(2016)]{Schoenberger_2016_sfm_colmap}
Johannes~Lutz Sch\"{o}nberger and Jan-Michael Frahm.
\newblock Structure-from-motion revisited.
\newblock In \emph{CVPR}, 2016.

\bibitem[Sch\"{o}nberger et~al.(2016)Sch\"{o}nberger, Zheng, Pollefeys, and Frahm]{Schoenberger_2016_mvs_colmap}
Johannes~Lutz Sch\"{o}nberger, Enliang Zheng, Marc Pollefeys, and Jan-Michael Frahm.
\newblock Pixelwise view selection for unstructured multi-view stereo.
\newblock In \emph{ECCV}, 2016.

\bibitem[Sch{\"o}nberger et~al.(2017)Sch{\"o}nberger, Price, Sattler, Frahm, and Pollefeys]{Schonberger_2016_retrieval}
Johannes~L. Sch{\"o}nberger, True Price, Torsten Sattler, Jan-Michael Frahm, and Marc Pollefeys.
\newblock A vote-and-verify strategy for fast spatial verification in image retrieval.
\newblock In \emph{Computer Vision -- ACCV 2016}, pages 321--337, Cham, 2017. Springer International Publishing.

\bibitem[Sergio~Izquierdo(2024)]{Izquierdo_2024_cliqueM}
Javier~Civera Sergio~Izquierdo.
\newblock Close, but not there: Boosting geographic distance sensitivity in visual place recognition.
\newblock In \emph{European Conference on Computer Vision (ECCV)}, 2024.

\bibitem[Sun et~al.(2017)Sun, Xie, Luo, and Wang]{Sun_2017_Baidu_dataset}
Xun Sun, Yuanfan Xie, Peiwen Luo, and Liang Wang.
\newblock A dataset for benchmarking image-based localization.
\newblock \emph{2017 IEEE Conference on Computer Vision and Pattern Recognition (CVPR)}, pages 5641--5649, 2017.

\bibitem[Taira et~al.(2018)Taira, Okutomi, Sattler, Cimpoi, Pollefeys, Sivic, Pajdla, and Torii]{Taira_2018_inloc}
Hajime Taira, Masatoshi Okutomi, Torsten Sattler, Mircea Cimpoi, Marc Pollefeys, Josef Sivic, Tomas Pajdla, and Akihiko Torii.
\newblock {InLoc}: Indoor visual localization with dense matching and view synthesis.
\newblock In \emph{IEEE Conference on Computer Vision and Pattern Recognition}, 2018.

\bibitem[{Torii} et~al.(2018){Torii}, {Arandjelović}, {Sivic}, {Okutomi}, and {Pajdla}]{Torii_2018_tokyo247}
A. {Torii}, R. {Arandjelović}, J. {Sivic}, M. {Okutomi}, and T. {Pajdla}.
\newblock 24/7 place recognition by view synthesis.
\newblock \emph{IEEE Transactions on Pattern Analysis and Machine Intelligence}, 40\penalty0 (2):\penalty0 257--271, 2018.

\bibitem[Tung et~al.(2024)Tung, Chou, Cai, Yang, Zhang, Wetzstein, Hariharan, and Snavely]{Tung_2024_Megascenes}
Joseph Tung, Gene Chou, Ruojin Cai, Guandao Yang, Kai Zhang, Gordon Wetzstein, Bharath Hariharan, and Noah Snavely.
\newblock Megascenes: Scene-level view synthesis at scale.
\newblock In \emph{ECCV}, 2024.

\bibitem[Wang et~al.(2019)Wang, Han, Huang, Dong, and Scott]{Wang_2019_multi_similarity_loss}
Xun Wang, Xintong Han, Weilin Huang, Dengke Dong, and Matthew~R Scott.
\newblock Multi-similarity loss with general pair weighting for deep metric learning.
\newblock In \emph{Proceedings of the IEEE Conference on Computer Vision and Pattern Recognition}, pages 5022--5030, 2019.

\bibitem[Warburg et~al.(2020)Warburg, Hauberg, López-Antequera, Gargallo, Kuang, and Civera]{Warburg_2020_msls}
Frederik Warburg, Søren Hauberg, Manuel López-Antequera, Pau Gargallo, Yubin Kuang, and Javier Civera.
\newblock Mapillary street-level sequences: A dataset for lifelong place recognition.
\newblock In \emph{2020 IEEE/CVF Conference on Computer Vision and Pattern Recognition (CVPR)}, pages 2623--2632, 2020.

\bibitem[Weyand et~al.(2020)Weyand, Ara{\'u}jo, Cao, and Sim]{Weyand_2020_gldv2}
Tobias Weyand, A. Ara{\'u}jo, Bingyi Cao, and Jack Sim.
\newblock Google landmarks dataset v2 – a large-scale benchmark for instance-level recognition and retrieval.
\newblock In \emph{IEEE Conference on Computer Vision and Pattern Recognition}, pages 2572--2581, 2020.

\end{thebibliography}
}


\end{document}